\documentclass[runningheads]{llncs}

\usepackage{comment}
\usepackage{color}

\usepackage{microtype}
\usepackage{graphicx}
\usepackage{xspace}
\usepackage[table]{xcolor}
\usepackage{pifont}%
\usepackage[T1]{fontenc}
\usepackage{lmodern}

\usepackage{epsfig}
\usepackage{graphicx}
\usepackage{wrapfig}
\usepackage{subcaption}
\usepackage[export]{adjustbox}

\usepackage{amsmath, amsthm, amssymb}
\usepackage{textcomp}
\usepackage{stmaryrd}
\usepackage{upgreek}
\usepackage{bm}
\usepackage{cases}
\usepackage{mathtools}
\usepackage{arydshln}
\usepackage{multirow}

\usepackage{paralist}
\usepackage{enumitem}

\usepackage{cite}

\usepackage{multirow}
\usepackage{rotating}
\usepackage{booktabs}
\usepackage{tabularx}
\usepackage{array}
\usepackage{makecell}

\usepackage{url}
\usepackage{xspace}
\usepackage[T1]{fontenc}
\usepackage{utils/mysymbols}

\newcommand{\mfc}{M4C\xspace}
\newcommand{\tvqa}{TextVQA\xspace}
\newcommand{\stvqa}{ST-VQA\xspace}

\newcommand{\spat}{spatially aware self-attention\xspace}
\newcommand{\SPAT}{Spatially Aware Self-Attention\xspace}
\newcommand{\SAM}{SA-M4C}

\newcommand{\cmark}{\ding{51}}%
\newcommand{\xmark}{\ding{55}}%
\newcommand{\rarrow}{\ding{213}}%

\def\R{\mathbb{R}}
\def\tx{\widetilde x}
\def\tvx{\mathbf{\widetilde x}}
\def\g{\mathcal{G}}
\def\eij{e_{i\rightarrow j}}
\def\GE{\mathcal{E}}

\def\vx{\mathbf{x}}
\def\vy{\mathbf{y}}
\def\vq{\mathbf{q}}
\def\vk{\mathbf{k}}
\def\vv{\mathbf{v}}
\def\tv{\mathcal{T}^x}
\def\te{\mathcal{T}^e}

\def\vya{\vy^{\text{ans}}}

\def\xq{X^{\text{ques}}}
\def\espa{\mathcal{E}_{\text{spa}}}
\def\cR{\mathcal{R}}

\usepackage[pagebackref=true,breaklinks=true,colorlinks,bookmarks=false]{hyperref}

\usepackage{tablefootnote}

\begin{document}

\pagestyle{headings}
\mainmatter
\def\ECCVSubNumber{946}  %

\title{Spatially Aware Multimodal \\Transformers for TextVQA} %

\author{
    Yash Kant\inst{1} \and
    Dhruv Batra\inst{1,2} \and
    Peter Anderson\inst{1}\thanks{Now at Google} \and \\
    Alexander Schwing\inst{3} \and 
    Devi Parikh\inst{1,2} \and 
    Jiasen Lu\inst{1} \thanks{Work was partially done as a member of PRIOR @ Allen Institute for AI} \and
    Harsh Agrawal \inst{1}
}

\authorrunning{Y. Kant et al.}
\titlerunning{SA-M4C for TextVQA}
\institute{
Georgia Institute of Technology \and
Facebook AI Research (FAIR) \and 
University of Illinois, Urbana-Champaign
}

\maketitle
\begin{abstract}
\vspace{-0.15in}
Textual cues are %
essential for everyday tasks like buying groceries and using public transport. To develop this assistive technology, we study %
the TextVQA task, i.e., reasoning about text in images to answer a question. %
Existing approaches are limited in their  use of spatial relations and rely on fully-connected transformer-based architectures to implicitly learn the spatial structure of a scene. 
In contrast, we propose a novel \spat layer such that each visual entity only looks 
at neighboring entities defined by a spatial graph. Further, each head in our multi-head self-attention layer focuses on a different subset of relations. Our approach has two advantages: (1) each head %
considers local context instead of dispersing the attention amongst all visual entities; (2) %
we avoid learning redundant features. We show that %
our model improves the absolute accuracy of current state-of-the-art methods on TextVQA by 2.2\% overall over an improved baseline, and 4.62\% on questions  that involve spatial reasoning and can be answered correctly using OCR tokens. Similarly on ST-VQA, we improve the absolute accuracy by 4.2\%. We further show that \spat improves visual grounding. 
\keywords{VQA, TextVQA, Self-attention}
\end{abstract}
\vspace{-0.15in}
\vspace{-5pt}
\section{Introduction}
\label{sec:intro}
\vspace{-5pt}

The promise of assisting visually-impaired users gives us a compelling reason to study Visual Question Answering (VQA)~\cite{antol2015vqa} tasks. A dominant class of questions ($\sim$20\%) asked by visually-impaired users on images of their surroundings involves reading text in the image~\cite{Bigham2010VizWizNR}. Naturally, the ability to reason about text in the image to answer questions such as ``Is this medicine going to expire?'', ``Where is this bus going?'' is of paramount importance for these systems.
To benchmark a model's capability to reason about text in the image, new datasets \cite{Singh2019TowardsVM, STVQA, OCRVQA} have been introduced for the task of Text Visual Question Answering (TextVQA).

\begin{figure*}[t]
    \begin{center}
        \includegraphics[width=0.93\linewidth]{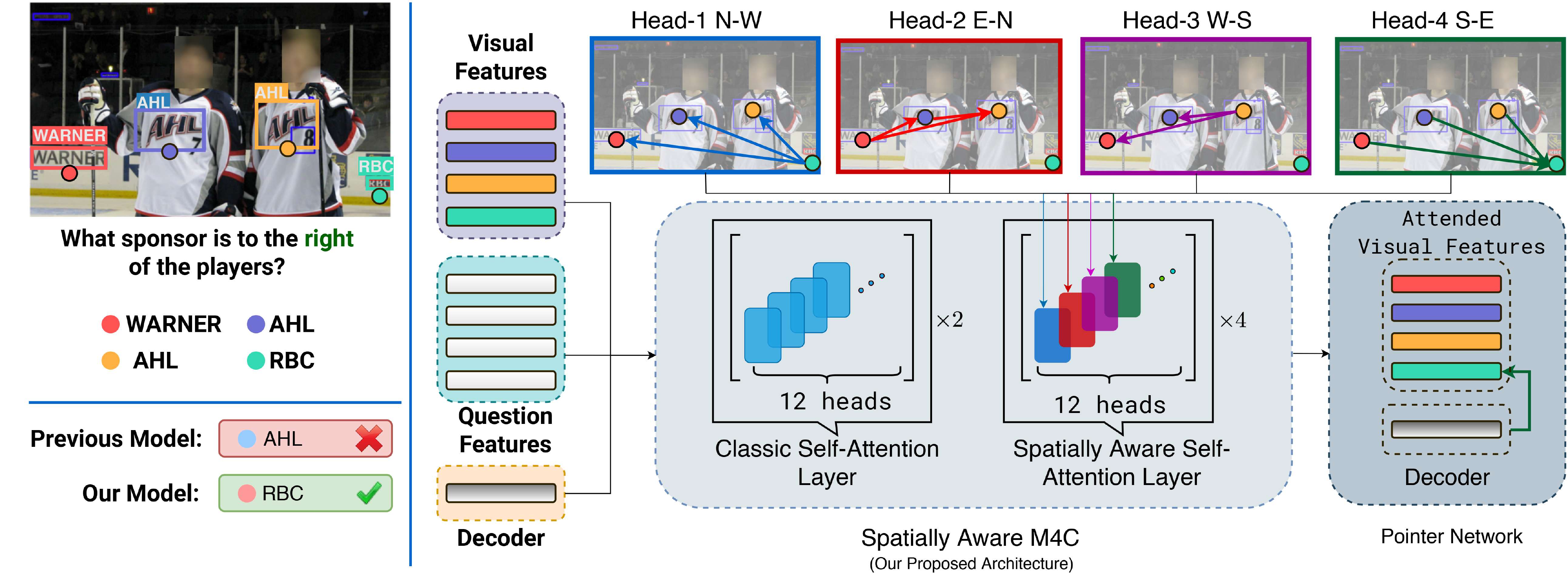}
        \label{fig: nocap_vs_coco_b}
    \end{center}
    \vspace{-15pt}
    \caption{\footnotesize
        (a) On questions that require spatial reasoning ($\sim$13\% of the TextVQA dataset),
        compared to previous approaches \cite{hu2019iterative,Singh2019TowardsVM},
        our model can reason about spatial relations between visual entities to answer questions correctly.
        (b) We construct a spatial-graph that encodes different spatial relationships between a pair of visual
        entities and use it to guide the self-attention layers present in multi-modal transformer architectures.
    }
    \vspace{-15pt}
    \label{fig:teaser}
\end{figure*}

Answering questions involving text in an image often requires reasoning about the relative spatial positions of objects and text. For instance, many questions such as ``What is written on the player's jersey?'' or ``What is the next destination for the bus?'' ask about text associated with a particular visual object. Similarly, the question asked in \figref{fig:teaser}, ``What sponsor is to the right of the players?'', explicitly asks the answerer to look to the \texttt{right} of the players. Unsurprisingly, $\sim$13\% of the questions in the TextVQA dataset use one or more spatial prepositions\footnote{We use several prepositions such as `right', `top', `contains', \etc to filter questions that involve spatial reasoning.}.

Existing methods for TextVQA reason jointly over 3 modalities -- the input question, the visual content and the text in the image. 
LoRRA~\cite{Singh2019TowardsVM} uses an off-the-shelf Optical Character Recognition (OCR) system~\cite{Borisyuk2018} to detect OCR tokens and extends previous VQA models~\cite{anderson2018bottom} to select single OCR tokens from the images as answers. The more recently proposed Multimodal Multi-Copy Mesh (\mfc) model~\cite{hu2019iterative} captures intra- and inter-modality interactions over all the inputs -- question words, visual objects and OCR tokens -- by using a multimodal transformer architecture that iteratively decodes the answer by choosing words from either the OCR tokens or some fixed vocabulary. The superior performance of \mfc is attributed to the use of multi-head self-attention layers~\cite{Vaswani2017AttentionIA} which has become the defacto standard for modeling vision and language tasks~\cite{ViLBERT, MultiTask, VLBert, Uniter, Lxmert}.

While these approaches take advantage of detected text, they are limited in how they use spatial relations. %
For instance, LoRRA~\cite{Singh2019TowardsVM} does not use any location information while \mfc~\cite{hu2019iterative} merely encodes the absolute location of objects and text as input to the model. By default, self-attention layers are fully-connected, dispersing attention across the entire global context and disregarding the importance of %
the local context around a certain object or text. As a result, in existing models the onus is on them to \textit{implicitly} learn to reason about the relative spatial relations between objects and text. 
In contrast, in the Natural Language Processing community, it has proven beneficial to \textit{explicitly} encode semantic structure between input tokens~\cite{SelfAttentionSPR, AdaptiveAST, Yang2018ModelingLF}. Moreover, while multiple independent heads in self-attention layers model different context, each head independently looks at the same global context and learns redundant features~\cite{MultiTask} that can be pruned away without substantially harming a model's performance~\cite{Voita2019AnalyzingMS,Michel2019AreSH}.

We address the above limitations by proposing a novel %
\spat layer for multimodal transformers. First, we follow~\cite{Yao2018ExploringVR,Li2019RelationawareGA} to build a spatial graph to represent relative spatial relations between all visual entities, i.e., all objects and OCR tokens. We then use this spatial graph to guide the self-attention layers in the multimodal transformer. We modify the attention computation in each head such that each entity attends to just the neighboring entities as defined by the spatial graph, and we restrict each head to only look at a subset of relations which prevents learning of redundant features.  

Empirically, we evaluate the efficacy of our proposed approach on the challenging TextVQA~\cite{Singh2019TowardsVM} and Scene-Text VQA (ST-VQA)~\cite{STVQA} datasets. We first improve the absolute accuracy of the baseline \mfc model on TextVQA by 3.4\% with improved features and hyperparameter optimization. We then show that replacing the fully-connected self-attention layers in the \mfc model with our spatially aware self-attention layers improves absolute accuracy by a further 2.2\% (or 4.62\% for the $\sim$14\% of TextVQA questions that include spatial prepositions and has a majority answer in OCR tokens). On ST-VQA our final model achieves an absolute 4.2\% improvement in Average Normalized Levenshtein Similarity (ANLS). 
Finally, we show that our model is more visually grounded as it picks the correct answer from the list of OCR tokens 8.8\% more often than \mfc.

\vspace{-3pt}
\section{Related Work}
\label{sec:rel}
\vspace{-3pt}

\noindent\textbf{Models for TextVQA:}   
Several datasets and methods~\cite{Singh2019TowardsVM, hu2019iterative, STVQA, OCRVQA, OCRVQA} 
have been proposed for the TextVQA task -- i.e., answering questions which require models to explicitly reason about text present in the image.
LoRRA~\cite{Singh2019TowardsVM} extends Pythia~\cite{jiang2018pythia} with an OCR attention branch to reason over a combined list of answers from a static vocabulary and detected OCR tokens. Several other models have taken similar
approaches to augmenting existing VQA models with OCR inputs~\cite{STVQA, OCRVQA, ICDAR2019}.
Building on the success of transformers~\cite{Vaswani2017AttentionIA} and BERT~\cite{Devlin2019BERTPO}, the Multimodal Multi-Copy Mesh (M4C) model~\cite{hu2019iterative} (which serves as our baseline) uses a multimodal transformer to jointly encode the question, image and text and employs an auto-regressive decoding mechanism to perform multi-step answer decoding. 
However, these methods are limited in how they leverage the relative spatial relations between visual entities such as objects and OCR tokens. Specifically, early models~\cite{STVQA, OCRVQA, ICDAR2019} proposed for the  TextVQA task did not encode any explicit spatial information while M4C~\cite{hu2019iterative} simply adds a location embedding of the absolute location to the input feature. We improve the performance of these models by proposing a general framework to effectively utilize the relative spatial structure between visual entities within the transformer architecture.

\noindent \textbf{Multimodal representation learning for Vision and Language:} 
Recently, several general architectures for vision and language~\cite{MultiTask, Uniter,UnicoderVL,VLBert,Lxmert,ViLBERT,
li2019visualbert, murahari2019large, Qi2020IMageBERTCP, 
sun2019videobert, Zhou2019UnifiedVP, hu2019iterative} were proposed that  reduce architectural differences across  tasks. These models (including M4C) typically fuse vision and language modalities by applying either self-attention~\cite{BahdanauCB14} or co-attention~\cite{lu2016hierarchical} mechanisms to capture intra- and inter-modality interactions. They achieve superior performance on many vision and language tasks due to their strong representation power and their ability to pre-train visual grounding in a self-supervised manner. Similar to \mfc, these methods add a location embedding to their inputs, but do not explicitly encode relative spatial information (which is crucial for visual reasoning). Our work takes the first step towards modeling relative spatial locations within the multimodal transformer architecture.

\noindent \textbf{Leveraging explicit relationships for Visual Reasoning:}  Prior work has used Graph Convolutional Nets (GCN)~\cite{Kipf2016SemiSupervisedCW} and Graph Attention Networks (GAT)~\cite{velickovic2018graph} to leverage explicit relations for image captioning~\cite{Yao2018ExploringVR} and VQA~\cite{Li2019RelationawareGA}. Both these methods construct a spatial and semantic graph to relate different objects. Although our relative spatial relations are inspired from~\cite{Yao2018ExploringVR, Li2019RelationawareGA}, our encoding  differs greatly.  First, \cite{Yao2018ExploringVR, Li2019RelationawareGA} looks at all the spatial relations in every attention head, whereas each self attention head in our model looks at different subset of the relations, i.e., each head is only responsible for a certain number of relations. This important distinction prevents spreading of attention over the entire global context and reduces redundancy amongst multiple heads.

\noindent \textbf{Context aware transformers for Language Modeling:} Related to the use of spatial structure for visual reasoning tasks, there has been a body of work on modeling the underlying structure in input sequences for language modeling tasks. Previous approaches have considered encoding the relative position difference between sentence tokens~\cite{Shaw2018SelfAttentionWR} as well as encoding the depth of each word in a parse tree and the distance between word pairs~\cite{SelfAttentionSPR}.
Other approaches learn to adapt the attention span for each attention head~\cite{AdaptiveAST, Yang2018ModelingLF}, rather than explicitly modeling context for attention. While these methods work well for sequential input like natural language sentences, they cannot be directly applied to our task since our visual representations are non-sequential.

\vspace{-5pt}
\section{Background: Multimodal Transformers}
\vspace{-10pt}
\label{sec:background}
Following the success of transformer~\cite{Vaswani2017AttentionIA} and BERT~\cite{Devlin2019BERTPO} based architectures on language modeling and sequence-to-sequence tasks, multi-modal transformer-style models~\cite{hu2019iterative,MultiTask, Uniter,UnicoderVL,VLBert,Lxmert,ViLBERT,li2019visualbert, murahari2019large, Qi2020IMageBERTCP, sun2019videobert, Zhou2019UnifiedVP} have shown impressive results on several vision-and-language tasks. Instead of using a single input modality (i.e., text), multiple
modalities are encoded as a sequence of input tokens and appended together to form  a single input sequence. Additionally, a type embedding unique to each modality
is added to distinguish amongst input token of different modalities. 

The core building block of the transformer architecture is a self-attention layer followed by a feed-forward network. The self-attention layer aims at capturing the direct relationships between the different input tokens.
In this section, we first briefly recap the
attention computation in the multi-head self-attention layer of the transformer and 
highlight some issues with classical self-attention layers.

\subsection{Self-Attention Layer}
\label{sec:self-attn}
A self-attention (SA) layer operates on an input sequence
represented by $N$ $d_x$-dimensional vectors $X = (\vx_1,\dots,\vx_N) \in \R^{d_x \times N}$ and computes  the attended sequence 
$\Tilde X= (\tvx_1 ,\dots,\tvx_N) \in \R^{d_x\times N}$.
For this, self-attention employs $h$ independent attention heads and applies the attention 
mechanism of Bahdanau \etal~\cite{BahdanauCB14} to its own input. 
Each head in a self-attention layer transforms the input sequence ${X}$ into 
query $Q^h=[\vq^h_1,\dots,\vq^h_N] \in \R^{d_h\times N}$, 
key $K^h=[\vk^h_1,\dots,\vk^h_N] \in \R^{d_h\times N}$, 
and value $V=[\vv^h_1,\dots,\vv^h_N] \in \R^{d_h\times N}$ vectors 
via learnable linear projections parameterized by 
$W^h_Q, W^h_K, W^h_V \in \R^{d_x \times d_h}$:
$$
    (\vq^h_i, \vk^h_i, \vv^h_i) =  (\vx_iW^h_Q, \vx_iW^h_K, \vx_iW^h_V) \quad\forall i\in[1, \dots, N].  
$$
Generally, $d_h$ is  set to $d_x/H$. Each attended sequence element $\tvx^h_i$ is then  computed via a 
weighted sum of value vectors, i.e., 
\begin{equation}
    \tvx^h_i = \sum_{j=1}^{n}\alpha^h_{ij}\vv^h_j.
    \label{eq:1}
\end{equation}
The weight coefficient $\alpha^h_{ij}$ is computed via a Softmax  over 
a compatibility function that compares the query vector $\vq^h_i$ with key vectors of
all the input tokens $\vk^h_j$, $j\in[1, \dots, N]$:
\begin{equation}
    \alpha_{ij} = \text{Softmax}\left(\frac{\vq^h_{i}(\vk^h_j)^T}{\sqrt{d_h}}\right). 
    \label{eq:2}
\end{equation}
The computation in Eq.~\eqref{eq:1} and Eq.~\eqref{eq:2} can be more compactly written as:
\begin{equation}
    \texttt{head}_h =  \mathcal{A}^h(Q^h,K^h,V^h) = \text{Softmax}\left(\frac{Q^h(K^h)^T}{\sqrt{d_h}}\right)V^h  \quad\forall h=[1,\dots,H].
    \label{eq:3}
\end{equation}
The output of all heads 
are then concatenated followed by a linear transformation with weights 
$W^O \in \R^{(d_{h}\cdot H) \times d_{x}}$. Therefore, in the case of multi-head 
attention, we obtain the attended sequence $\Tilde X = (\tx_i, \dots, \tx_N)$ from
\begin{equation}
    \Tilde X =  \mathcal{A}(Q, K, V) = \left[\texttt{head}_1,\dots,\texttt{head}_H\right]W^O.
    \label{eq:4}
\end{equation}

\noindent \textbf{Application to multi-modal tasks}: For multi-modal 
tasks, the self-attention is often modified to model cross-attention from one modality
$U_i$ to another modality $U_j$ as $\mathcal{A}(Q_{U_i}, K_{U_j}, V_{U_j})$ or 
intra-modality attention $\mathcal{A}(Q_{U_i}, K_{U_i}, V_{U_i})$. Note, $U_i, U_j$ are simply sets of indices which are used to construct sub-matrices. Some architectures 
like \mfc \cite{hu2019iterative} use the classical self-attention layer to model attention between 
tokens of all the modalities as $\mathcal{A}(Q_{U}, K_{U}, V_{U})$ where
$U = U_1 \cup U_2 \cup \dots \cup U_{M}$ is the union of all $M$ input modalities.

\subsection{Limitations}
The aforementioned self-attention layer exposes two limitations: 
(1) self-attention layers model the global context by encoding 
relations between every single pair of input tokens. This disperses
the  attention across every input token and overlooks the importance
of semantic structure  in the sequence. For instance, in the 
case of language modeling, it has proven beneficial to capture local-context \cite{Yang2018ModelingLF} or the hierarchical
structure of the input sentence by encoding the depth of each word in a parsing tree~\cite{SelfAttentionSPR},
(2) multiple heads allow self-attention layers to jointly attend to different
context in different heads. However, each head independently looks at 
the entire global information and there is no explicit mechanism to ensure that 
different attention heads capture different context. Indeed, it has been 
shown that  the heads can be pruned away without substantially hurting a model's 
performance~\cite{Voita2019AnalyzingMS,Michel2019AreSH} and that different heads learn
redundant features~\cite{MultiTask}.

\vspace{-5pt}
\section{Approach}
\label{sec:approach}
\vspace{-5pt}

To address both limitations, we extend the self-attention layer
to utilize a graph over the input tokens. Instead of looking at the entire global context, an entity attends to just the neighboring entities as defined by a relationship graph. Moreover, heads consider different types of relations which encodes
different context and avoids learning redundant features. In  what follows, we introduce the notation for input token representations. Next, we formally define the heterogeneous graph over tokens from multiple modalities which are connected by different edge types. Finally, we describe our approach to adapt the attention span of each head in the self-attention layer by utilizing this graph. While our framework is general and easily extensible to other tasks, we present our approach for the \tvqa task.

\subsection{Graph over Input Tokens}
\label{sec:general-graph} Let us define a directed cyclic heterogeneous graph $\g = (X, \GE)$ where each node corresponds to an 
input token $\vx_i \in X$. $\GE$ is a set of all edges $\eij,  \forall \vx_i,\vx_j\in X$. Additionally, we define a mapping function $\Phi_x: X \to \tv$ 
that maps a node $\vx_i \in X$ to one of the modalities. Consequently the number of node types is
equal to the number of input modalities, i.e., $|\tv| = M$. We also define a mapping function $\Phi_e: \mathcal{E} \to \te$ that maps an edge $\eij \in \mathcal{E}$ to a  
relationship type $t_l \in \mathcal{T}^e.$ 

We represent the question as a set of tokens, i.e.,   
$X^\text{ques} = \{\vx \in X: \Phi_x(\vx) = \text{ques}\}$. 
The visual content in the image is represented via a list of  object region features $X^\text{obj} = \{\vx \in X: \Phi_x(\vx) = \text{obj}\}$. Similarly, the list of OCR tokens present in the image is referred to as 
$X^\text{ocr} = \{\vx \in X: \Phi_x(\vx) = \text{ocr}\}$. Following \mfc, the model decodes multi-word answer $Y^{\text{ans}} = (\vya_1, \dots, \vya_T)$ for $T$ time-steps.\smallskip
\begin{figure*}[t]
    \begin{center}
        \includegraphics[width=0.80\linewidth]{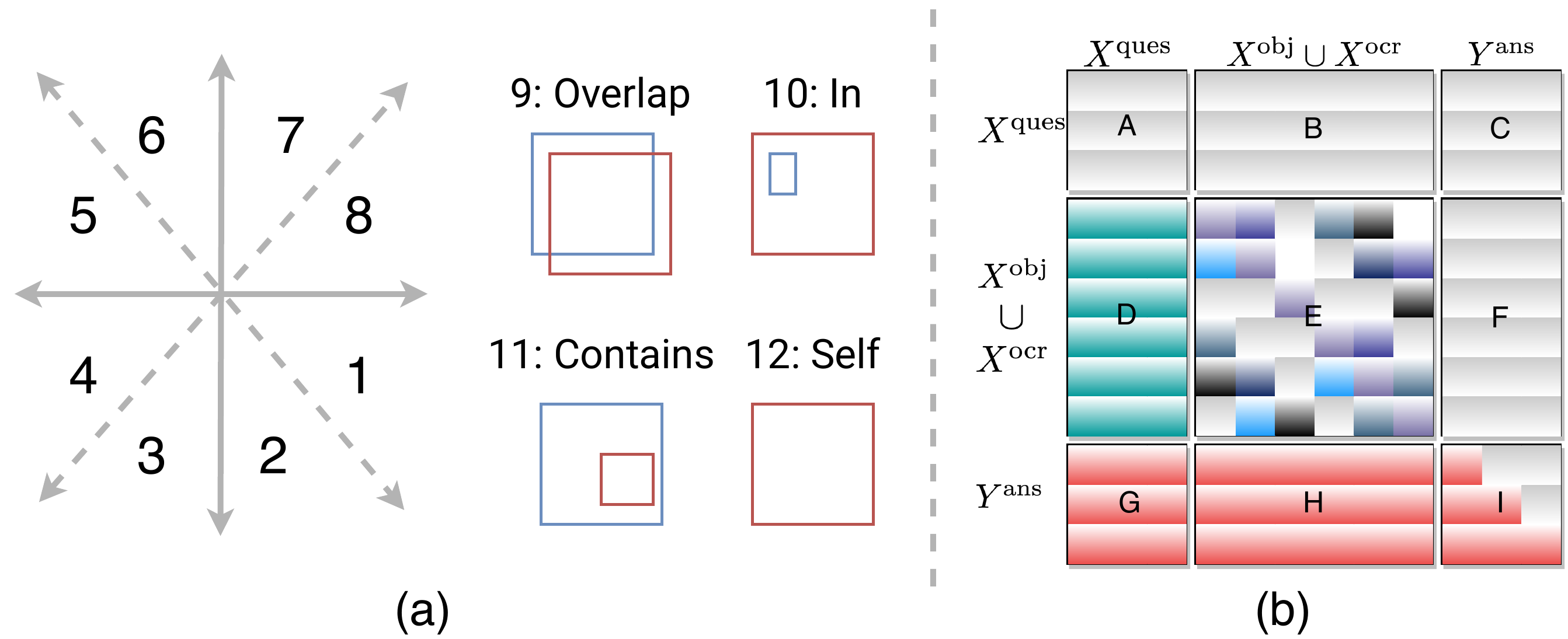}
    \end{center}
    \vspace{-15pt}
    \caption{
        \footnotesize{(a) The spatial-relations graph encodes twelve type of relations between two object or OCR tokens $r_i, r_j \in \mathcal{R}$. (b) Denotes the attention mask between different data modalities. In our \spat~layer, object and OCR tokens attend to each other based on a subset of spatial relations $\mathcal{T}^h \subseteq \mathcal{T}^{\text{spa}}$. They also attend to question tokens via $t_{\text{imp}}$ relation. Any input token $x\in X$ do not attend to answer token $y^{\text{ans}} \in Y$ while $y^{\text{ans}}$ can attend to tokens in $X$ as well as previous answer tokens $y^{\text{ans}}_{<t}$.
    }}
    \label{fig:spatial-relations}
    \vspace{-15pt}
\end{figure*}
\noindent \textbf{Spatial Relationship Graph}:
Answering questions about text in the image involves reasoning about the spatial relations between various OCR tokens and objects present in the image.
For instance, the question ``What is the letter on the player's hat?''  requires to first detect a hat in the image and then reason about the `\texttt{contains}' relationship between the letter and the player's hat. 

To encode these spatial relationships between all the
objects $X^\text{obj}$ and OCR tokens $X^\text{ocr}$ present in the image, i.e., all the regions $r\in\cR = X^\text{obj}\cup X^\text{ocr}$, we construct a spatial graph $G_{\text{spa}} = (\cR, \espa)$ with nodes corresponding to the union of all objects and OCR tokens. The mapping function $\Phi_{spa}: \espa \to \mathcal{T}^{\text{spa}}$ assigns a spatial relationship $t_l \in \mathcal{T}^{\text{spa}}$ to  an edge $e = (r_i,r_j)\in\espa$.
The mapping function utilizes the rules introduced by Yao \etal~\cite{Yao2018ExploringVR} which we illustrate in~\figref{fig:spatial-relations}(a). 
We use a total of twelve types of spatial relations (e.g., $\langle r_i-\texttt{contains}-r_j\rangle$, $\langle r_i-\texttt{is-inside}-r_j\rangle$ as well as a `$\texttt{self-relation}$'). Note that $G_{\text{spa}}$ is a symmetric directed
graph, i.e., for every edge $\eij$ there is a reverse edge $e_{j\to i}$.\smallskip

\noindent \textbf{Implicit Relationship between Objects, OCR and Question Tokens}: For the TextVQA task, different types of spatial relations might be useful for different question types. For instance, a question asking about `what is written on the player's jersey' might focus on the \texttt{contains} relationship, whereas a question asking about `what sponsor is to the right of the player' might utilize the \texttt{right} relationship. Thus, to inject semantic information from the question into the object and OCR representation, we allow object and OCR tokens to attend to question tokens.
In our general framework, we  accomplish this via a bipartite graph $G_{\text{imp}}(\cR, \xq, \mathcal{E}_\text{imp})$ connecting all the object and OCR tokens $r_i\in\cR$ to all question tokens $ \vx_j \in X^{\text{ques}}$ via an implicit edge $\eij$ of type $t_{\text{imp}}$. Thus, by attending to question tokens, each object and OCR token learns to implicitly incorporate useful semantic information from the question into its representation. 

\begin{figure*}[t]
    \begin{center}
        \includegraphics[width=0.93\linewidth]{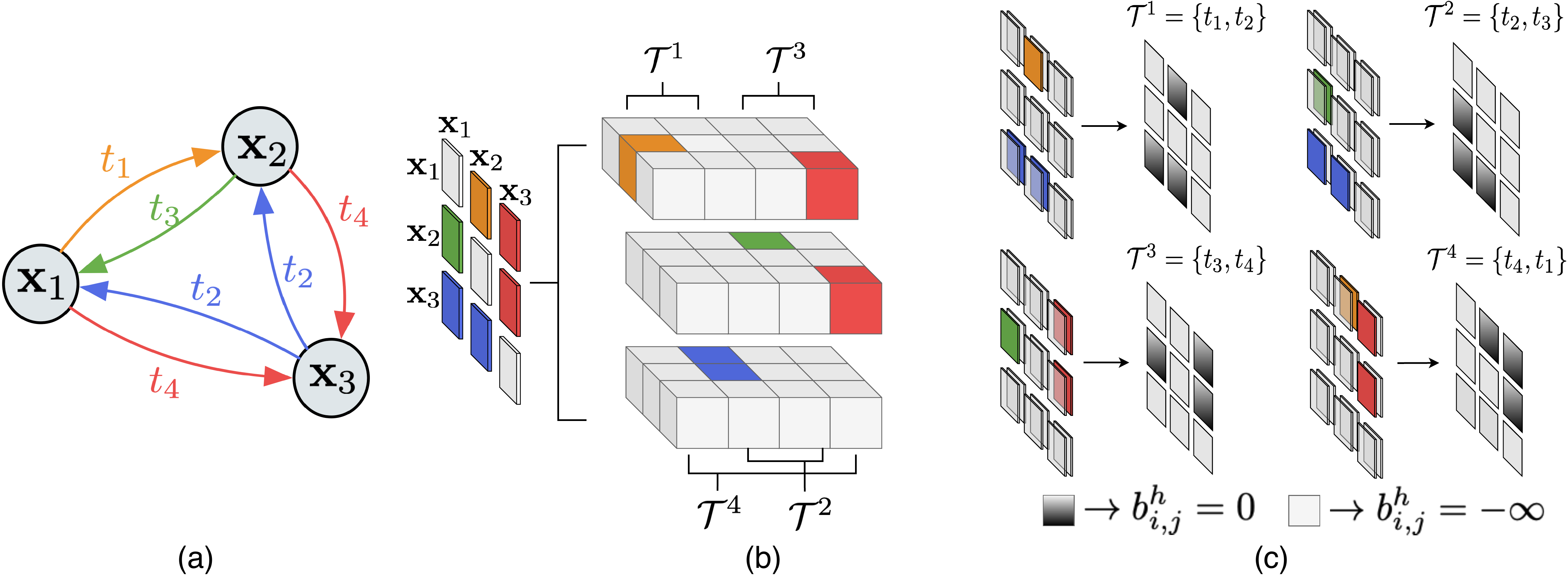}
        \label{fig: nocap_vs_coco_b}
    \end{center}
    \vspace{-20pt}
    \caption{\small
        (a) Spatially aware attention layer uses a spatial graph to guide the attention in each head of the self-attention layer.
        (b) The spatial graph is represented as a stack of adjacency matrices, each for a given relationship $t_e$ 
        (c) Each head indexed by $h$ looks at a subset of relationships $\mathcal{T}^h$ defined by the size of the context ($c=2$ here), e.g. $\texttt{head}_1$  looks at a two types of relation ($\mathcal{T}^1 = \{t_1, t_2\}$). When edge $\eij \in \mathcal{T}^h$ (black box), bias term is set to $\beta^{h}_{i,j}=0$, otherwise when $\eij \notin \mathcal{T}^h$ (white blocks), $\eij \notin \mathcal{T}^h = -\infty.$
    } 
    \label{fig:graph_aware_sa}
    \vspace{-16pt}

\end{figure*}
\subsection{\SPAT Layer}
\label{sec:gasa}

As mentioned in Section~\ref{sec:background}, attention in a 
single-head $h$ of a self-attention layer
can be computed by the compatibility function defined in Equation~\eqref{eq:2}. The compatibility function computes a similarity  between a  query $\vq_i^h$ corresponding to input $\vx_i$, and the key vector $\vk_j^h$ of input token $\vx_j$. Within a single head we want attention to only look at relevant tokens. We model it by allowing each head to focus on only a subset of edge types $\mathcal{T}^{h} \subseteq \mathcal{T}^{e}$. In other words, we want each token $\vx_i$ to only focus on tokens $\vx_j$ when they are connected via an edge $\eij$ of type $\Phi_e(\eij) \in \mathcal{T}^h$. \smallskip

In the context of \tvqa, we use the combination of two graphs, $G_{\text{spa}} \cup G_{\text{imp}}$, defined over tokens from all
the input data modalities $\vx \in X$. The subset of 
relations $\mathcal{T}^{h}$ each head $h$ attends to  is subset of $c$ spatial relationships between $(\vx_i, \vx_j)$ and one implicit relationship between question and image tokens, i.e., 
$$ \mathcal{T}^{h} = \{t_{\text{imp}}, t_h, t_{h+1}, \cdots, t_{(h+c)\bmod |\mathcal{T}^{\text{spa}}|}\}, \quad t \in \mathcal{T}^e =  \mathcal{T}^\text{spa}\cup t_\text{imp}.$$

When $c > 1$, multiple heads are aware of a given spatial relationship and we are encouraging 
the models to jointly attend to information from different representation subspaces~\cite{Vaswani2017AttentionIA}. When $c=1$, each head only focuses on a one type of spatial relationship. 
Similarly when $c=|\mathcal{T}^{spa}| + 1$, each head attends to all the spatial relationships
as well as the implicit relationship $t_{\text{imp}}$. We empiricaly observed that $c=2$ works best for our setting.

As illustrated in~\figref{fig:graph_aware_sa}, to weigh the attention in each head based on the subset of spatial relationships $\mathcal{T}^h$, we introduce a bias term defined as 
\begin{equation}
    b^h_{i, j} = \begin{cases}
    \beta^h_{t_l} & t_l \in \mathcal{T}^h, \quad \vx_i, \vx_j \in X\\
    -\infty & \text{otherwise}
    \end{cases},
    \label{eq:5}
\end{equation} 
to modify the computation of the attention weights $\alpha^h_{ij}$ over different tokens. Specifically, we compute attention weights %
as follows\footnote{If for a given $\vx_i$, $\Phi_{e}(\eij) \notin \mathcal{T}^h,   \forall j\in [1,N]$, then we explicitly set $\alpha^h_{i,j}=0,$.}: 
\begin{equation}
    \alpha^h_{ij} = \text{Softmax}\left(\frac{\vq_{i}^{h}(\vk_j^{h})^{T} + b^h_{i,j} }{\sqrt{d_h}}\right).
    \label{eq:6}
\end{equation}
Intuitively, as illustrated in \figref{fig:graph_aware_sa} if there is no edge $\eij$ of type $t_l \in \mathcal{T}^h$ between nodes $\vx_i$ and $\vx_j$, then the compatibility score $\vq_{i}^{h}(\vk_j^{h})^{T} + b^h_{i,j}$ is negative infinity and the  attention weights $\alpha^h_{ij}$ become zero. Otherwise, the attention weights can be modulated based on the specific
edge type $t_l = \Phi_e(\eij)$ by learning a bias term for 
each edge type $\beta^h_{t_l} \in \{\beta^h_{t_1},  \dots, \beta^h_{|\mathcal{T}^e|}\}$. Alternatively, we can set $\beta^h_{t_l}$ to zero if we do not want to modulate attention based on the edge type between a pair of tokens. Classical self-attention layers described in Section~\ref{sec:self-attn} are hence a special case which is obtained when $|\mathcal{T}^e|=1$ and $G$ is a fully connected graph.\smallskip

Specifically, for \tvqa, as illustrated in~\figref{fig:spatial-relations}(b), all object and OCR tokens attend to each other based on that subset of relations $\mathcal{T}^h$ that a head is responsible for. Since we want the representations of object and OCR tokens to contain information about the question, all object and OCR tokens 
attend to all question tokens via the edge of type  $t_{\text{imp}}$. For simplicity, we don't learn this relation-dependent bias and run all our experiments with $\beta^h_{t_l}$ set to zero. 

Importantly, our graph-aware self attention layer overcomes the aforementioned two limitations of classical self-attention layers. First, each head is able to focus on a subset of relations $\mathcal{T}^h \subseteq \mathcal{T}^e$. Consequently, the attention is not distributed over the entire
sequence of tokens and each token gathers information from only a specific  subset of tokens. Second, we are forcing each head to look at a different context which prevents the heads from learning redundant features. \smallskip

\noindent \textbf{Causal Attention for Answer tokens}: During decoding, 
the \mfc model generates answer tokens $\vya_t$ $\forall t$ one step at a time. Inspired by the success of several text-to-text models~\cite{Liu2018GeneratingWB, Dong2019UnifiedLM}, the \mfc architecture uses a causal attention mask where $\vya_t$  attends to all question, image, and OCR tokens $\vx\in X$ along with entries in the answer $\vya_{<t}$ prior to time $t$. We follow \cite{hu2019iterative, Singh2019TowardsVM} to generate the answer tokens. During decoding, at each step the model transforms the predicted token from the previous step to a d-dimensional vector $z_t$. We use $z_t
$ to compute similarity with all OCR-tokens and vocabulary words and pick the the most similar one. We iteratively decode the answer over 12 time steps. 

\subsection{Implementation Details}
\vspace{-3pt}
Following \mfc\cite{hu2019iterative}, the input to our multimodal transformer consists of three different modalities -- 1) 20 Question tokens, 2) 100 Object tokens, and 3) 50 OCR tokens. Below, we briefly describe the construction of each of the modal features. For a detailed discussion we refer the reader to \mfc~\cite{hu2019iterative}.\smallskip\\
\textbf{Question Features:} We encode the question text using  three layers of a BERT~\cite{Devlin2019BERTPO} model pre-trained on English Wikipedia and Book-Corpus~\cite{Zhu2015AligningBA} datasets. We finetune this model during training.\smallskip\\
\textbf{Object Features:} We encode the object regions by extracting features from a ResNeXT-152~\cite{Xie_2017} based Faster R-CNN model~\cite{ren2015faster} trained on Visual Genome~\cite{Krishna2016VisualGC} with attribute loss. We then add an absolute-location embedding to these features by using the bounding box coordinates. \smallskip\\
\textbf{OCR Features:} Similarly, for OCR, we extract region features using the same object detector and we append an embedding obtained from FastText~\cite{bojanowski2017enriching} and PHOC features~\cite{AlmazanGFV14} of the ocr-text. We also add an absolute-location embedding by using the bounding box coordinates of the OCR token similar to object features.
\vspace{-3pt}
\section{Experiments}
\label{sec:exp}
We evaluate our model on the TextVQA dataset \cite{Singh2019TowardsVM} and the ST-VQA dataset \cite{STVQA}. Our model outperforms previous work by a significant margin and sets the new state-of-the-art on both datasets.

\subsection{Evaluation on TextVQA dataset}
The TextVQA dataset~\cite{Singh2019TowardsVM} contains 28,408 images from
the Open Images dataset~\cite{OpenImages}, with human-written questions
asking about text in the image. Following VQAv2~\cite{antol2015vqa}, each question in the TextVQA dataset has 10 free-response answers, and the final accuracy is measured via soft voting of the 10 answers (VQA Accuracy). Following the \mfc model~\cite{hu2019iterative}, we collect the top 5000 frequent words from the answers in the training set as our answer vocabulary.
We compare our method with the recent proposed LoRRA \cite{Singh2019TowardsVM}, M4C \cite{hu2019iterative} and 2019 TextVQA Challenge leaderboard entires \cite{textvqa_RTArt, textvqa_msft_vti}.

\noindent \textbf{Improved M4C baseline (M4C$^\dagger$).} To establish a strong baseline we further improve \mfc by replacing the Rosetta-en OCR system with the Google OCR system\footnote{\url{https://cloud.google.com/products/ai/}} which we qualitatively find to be more accurate, detecting text with higher recall and having fewer spelling errors. This improves the performance from 39.4\% to 41.8\% (Rows \texttt{4} and \texttt{6} in Table~\ref{tab:textvqa}). Next, we replace the ResNet-101~\cite{He_2016} backbone of the Faster R-CNN~\cite{ren2015faster} feature-extractor with a ResNeXt-151 backbone~\cite{Xie_2017} as recommended by~\cite{MultiTask}. This further improves the performance from 41.8\% to 42.0 \% (Rows \texttt{6} and \texttt{7}). Finally, we add two additional transformer layers (Row \texttt{8}), jointly train \mfc on \stvqa~\cite{STVQA} (Row \texttt{9}) and use beam search decoding (Row \texttt{10}) to establish the final improved baselines 43.8\% and 42.4\% on validation and test set respectively.

\newcommand{\band}{\rowcolor{gray!10}}
\begin{table*}[t!]\footnotesize
\setlength\tabcolsep{3.5pt}
\renewcommand{\arraystretch}{1.25}
\center
\caption{Results on TextVQA \cite{Singh2019TowardsVM} dataset. We compare our model (rows 11-13) against the prior works (row 1-5) and the improved baselines (rows 6-10).
\\ \scriptsize{$^{\dagger}$ Indicates our ablations for improved baseline.\\$^{\dagger \dagger}$ Indicates the best model from improved baseline.}}
\resizebox{0.73\textwidth}{!}{
\begin{tabular}{clccccccc}
\toprule
& Method & Structure & \makecell{OCR \\ system} & \makecell{DET \\backbone} & \makecell{w/ \\ ST-VQA} & \makecell{Beam \\ size} & \makecell{Accu. \\ on val} & \makecell{Accu. \\ on test} \\
\midrule
\band \small\texttt{1} & LoRRA \cite{Singh2019TowardsVM} & - & R-ml & ResNet & \xmark & - & 26.5 & 27.6\\
\small\texttt{2} & DCD \cite{textvqa_RTArt} & - & - & - & - & - & 31.4 & 31.4\\
\band \small\texttt{3} & MSFT \cite{textvqa_msft_vti} & - & - & - & - & - & 32.9 & 32.4\\
\small\texttt{4} & M4C \cite{hu2019iterative} & 4N & R-en & ResNet &  \xmark & 1 & 39.4 & 39.0 \\
\band \small\texttt{5} & M4C \cite{hu2019iterative} & 4N & R-en & ResNet &  \cmark & 1 & 40.5 & 40.4\\
\midrule
\small\texttt{6} & M4C \cite{hu2019iterative}$^{\dagger}$ & 4N & G & ResNet & \xmark & 1 & 41.8 & - \\
\band \small\texttt{7} & M4C \cite{hu2019iterative}$^{\dagger}$ & 4N & G & ResNeXt & \xmark& 1 & 42.0 & - \\
\small\texttt{8} & M4C \cite{hu2019iterative}$^{\dagger}$ & 6N & G & ResNeXt & \xmark& 1 & 42.7 & - \\
\band \small\texttt{9} &  M4C \cite{hu2019iterative}$^{\dagger}$ & 6N & G & ResNeXt & \cmark & 1 & 43.3 & - \\
\small\texttt{10} &  M4C \cite{hu2019iterative}$^{\dagger \dagger}$ & 6N & G & ResNeXt & \cmark & 5 & 43.8 & 42.4\\
\midrule
\band \small\texttt{11} & \SAM \ (ours) & 2N\rarrow4S & G & ResNeXt & \xmark & 1 & 43.9 & -\\
\small\texttt{12} & \SAM \ (ours) & 2N\rarrow4S & G & ResNeXt & \cmark & 1 & 45.1 & - \\
\band \small\texttt{13} & \SAM \ (ours) & 2N\rarrow4S & G & ResNeXt & \cmark & 5 & \textbf{45.4} & \textbf{44.6} \\
\bottomrule
\end{tabular}}
\smallskip
\label{tab:textvqa}
\vspace{-15pt}
\end{table*}

\noindent \textbf{Our Results (SA-M4C).} 
Our model consist of 2 normal self-attention layers and 4 \spat layers (2N\rarrow4S). As shown in Table.~\ref{tab:textvqa} Row \texttt{13}, our model is 2.2\% (absolute) higher than it's counterpart in Row \texttt{10} and 4.4\% better than the baseline \mfc model (Row \texttt{5}). Note that the improved \mfc model in Row \texttt{10} and our method use the same input features, equal number of transformer layers and have the same number of parameters. Next, we perform model ablations to analyze the source of the gains in our method.

\noindent \textbf{Model structure ablations.} We answer the question, "How many \spat layers are helpful?", by incrementally replacing the self-attention layers in \mfc with the proposed \spat\ layers. 
Table.~\ref{tab:ablations} Row \texttt{1}, \texttt{2} and \texttt{3} show that the performance improves as we replace normal self-attention layers with \spat. We achieve the best performance after replacing 4 out of 6 self-attention layers (43.19\% \vs 43.80\%). It's important to note that keeping the bottom self-attention layer is critical to model attention across modalities since attention for question tokens are masked in \spat.

\noindent \textbf{Span of \spat head.} 
Recall that the context-size parameters ($c$) is the number of relationships $|\mathcal{T}^h|$ each attention head looks at, and controls the sparsity of each head in \spat. When $c > 1$, multiple heads are aware of a given spatial relationships which jointly attend information from different representation subspaces. Sweeping over the context-size ($c$) we find that $c=2$ works the best (Row \texttt{7} in Table~\ref{tab:ablations}).

\noindent \textbf{Comparing with other methods that induce sparsity into transformer. } 
We further compare our approach with other formulations~\cite{Li2019RelationawareGA, ExplicitST} that induce sparsity in the Transformer architectures as well as randomly mask attention heads. We describe each setting as follows.
\begin{compactitem}[\hspace{3pt}--]
\item \textbf{Random masking (M4C-Random).} We randomly initialize a spatial graph by assigning an edge of a given type between two nodes including a no-edge with equal probability. We use this graph as input to our \spat layer. Through this comparison, we want to establish the importance of spatial graph induced sparsity vs random sparsity in self-attention layers. We report this baseline by averaging across 5 different seeds. 

\item \textbf{Top-k Attention (M4C-Top-k). } Instead of masking the attention weights based on a graph, we explicitly select the $top \mbox{-} k$ attention weights and mask the rest \cite{ExplicitST}. We use $k = 9$ which corresponds to inducing the same level of sparsity as our baseline model. This helps to establish the need of guiding the attention based on spatial relationships.

\item \textbf{Graph Attention (M4C-ReGAT). } We  implement ReGAT-based attention layer Li \etal~\cite{Li2019RelationawareGA} which endows Graph Attention Network~\cite{velickovic2018graph} encoding with spatial information by adding a bias term specific to each relation.\footnote{We use the code released by the authors \url{https://github.com/linjieli222/VQA_ReGAT/}}. The goal is to establish the improvements of our \spat layer compared to prior work.
\end{compactitem}

\noindent Table.~\ref{tab:ablations} Row \texttt{4} - Row \texttt{7} show these comparisons. We observe that random masking decreases the performance on TextVQA dataset by 1.2\% which verifies the importance of a correct spatial relationship graph. By selecting the $top \mbox{-} k$ connections (M4C-Top-9), we observe an improvement of 0.66\% compared to M4C-Random. However, M4C-Top-9 still underperforms compared to our proposed SA-M4C model by 0.64\%. Similarly, our proposed SA-M4C model outperforms the graph attention version (M4C-ReGAT) by 0.7\%.

\begin{table}[t]\footnotesize
\centering
\makebox[0pt][c]{\parbox{1.0\textwidth}{%
\begin{minipage}[b]{0.465\hsize}\centering
\caption{\small Model ablations on TextVQA.}
\resizebox{1\textwidth}{!}{
\begin{tabular}{clccc}
\toprule
& Method & Struc. & Context & Accu.(val) \\
\midrule
        \band \texttt{1} & \mfc\cite{hu2019iterative}$^{\dagger}$  & 6N & - & 42.70 \\
        \midrule
        \texttt{2} & \SAM \ (ours) & 4N\rarrow2S & 1 & 43.19 \\
        \band \texttt{3} & \SAM \ (ours) & 2N\rarrow4S & 1 & 43.80 \\
        \midrule
        \texttt{4} & \mfc-Random & 2N\rarrow4S & 1 & 42.09\\
        \band\texttt{5} & \mfc-Top-9 \cite{ExplicitST} & 2N\rarrow4T & - & 43.26\\
        \texttt{6} & \mfc-ReGAT \cite{Li2019RelationawareGA} & 2N\rarrow4Re & - & 43.20 \\
        \band\texttt{7} & \SAM \ (ours) & 2N\rarrow4S & 2 & \textbf{43.90} \\
        \bottomrule
\end{tabular}}
\label{tab:ablations}
\end{minipage}
\hfill
\begin{minipage}[b]{0.52\hsize}
\caption{\small Results on ST-VQA dataset.}
\resizebox{1\textwidth}{!}{
\begin{tabular}{clccccccc}
\toprule
& Method & Struc. & \makecell{Beam \\ size} & \makecell{VQA \\ Accu.} & \makecell{ANLS \\ on val} & \makecell{ANLS \\ on test} \\
\midrule
\band \texttt{1} & SAN+STR \cite{STVQA} & - & - & - & - & 0.135 \\
\texttt{2} & VTA \cite{ICDAR2019} & -& -& - & - & 0.282 \\
\band \texttt{3} & M4C \cite{hu2019iterative}   & 4N & 1 & 38.05 & 0.472 & 0.462 \\
\texttt{4} & \mfc\cite{hu2019iterative}$^\dagger$  & 6N &  1 & 40.71 & 0.499 & - \\
\midrule
\band\texttt{5} & \SAM \ (ours) & 2N\rarrow4S & 1  & 42.12  & 0.510  & - \\
\texttt{6} & \SAM \ (ours) & 2N\rarrow4S & 5  & \textbf{42.23}  & \textbf{0.512}  & \textbf{0.504} \\

\bottomrule
\end{tabular}}
\label{tab:stqa}
\end{minipage}
}}
\vspace{-18pt}
\end{table}

\subsection{Evaluation on ST-VQA}
We also report results on the \stvqa~\cite{STVQA} dataset which is another recently proposed dataset for the \tvqa task. \stvqa contains 18,921 training and  2,971 test images sourced from several datasets~\cite{ICDAR2013, Karatzas2015ICDAR2C, Bigham2010VizWizNR, MishraICCV13, Krishna2016VisualGC}. %
Following \mfc, we report results on the Open Dictionary (Task-3) as it matches the TextVQA setting where no answer candidates are provided at test time.

The ST-VQA dataset adopts Average Normalized Levenshtein Similarity (ANLS) defined as  $1 - d_L(a_\text{pred}, a_\text{gt}) / \max(|a_\text{pred}|, |a_\text{gt}|)$ averaged over all questions. $a_\text{pred}$ and $a_\text{gt}$ refer to prediction and ground-truth answers respectively while $d_L$ is edit distance. The metric truncates scores lower than 0.5 to 0 before averaging. We use both VQA accuracy and ANLS as the evaluation metric to facilitate comparison with prior work.

For training and validation on \stvqa we use the same splits used by \mfc~\cite{hu2019iterative} generated by randomly selecting 17,028 images for  training and the remaining 1,893  for validation. We train the improved baseline model and our best model (\spat) on \stvqa and report results in Table~\ref{tab:stqa}. Following prior works~\cite{STVQA, hu2019iterative} we show VQA Accuracy and ANLS both on validation set and only the latter on the test set. On the validation set our improved baseline achieves an accuracy of 40.71\% and an ANLS of 0.499 improving by 2.66\% and 0.027 absolute. Further, the final model with \spat\ layers achieves an accuracy of 42.23\% and an ANLS of 0.512 improving by 1.52\% and 0.013 in absolute gains on the  validation set. On the test set, our best model achieves state-of-the-art performance of of 0.504 ANLS.

\vspace{-5pt}
\section{Analysis}
\vspace{-5pt}

\textbf{Spatial reasoning}: We look at the source of improvements in our model both quantitatively and qualitatively. First, we look at the performance of our model on subset of questions from TextVQA validation dataset that involve spatial reasoning. For this, we carefully curate a list of spatial-prepositions (see Supplementary for detail),  and filter questions based on occurrence of one or more of these spatial-prepositions. After applying this filter, We observe that $\sim14\%$ of the questions (709/5000) are retained. On this subset ${D}_{\text{spa}}$, our model perform $2.83\%$ better than \mfc. Since, OCR tokens can answer only $\sim65\%$ of the questions in the validation set, we also look at the subset of questions that require spatial reasoning and has a majority answer in OCR tokens. On this subset $D_{\text{spa+ocr}}$ (409/5000 questions), our model performs $4.62\%$ better than \mfc.\smallskip

\noindent \textbf{Visual Grounding}: As a proxy to analyze visual grounding of our model, we look at instances in which models predict the answer using the list of OCR tokens without relying on the vocabulary. Our model picks an answer from the list of OCR tokens on 368/701 questions from the ${D}_{\text{spa}}$ subset, and achieves $52.85\%$ accuracy. This greatly improves the performance over \mfc which only achieves $44.05\%$ accuracy on a similar number (398/709) of questions that were answered using OCR tokens. The increase in performance is similar on $D_{\text{spa+ocr}}$ where we achieve a score of $67.95\%$ on 260/401 questions compared to $59.27\%$ achieved by \mfc over 273/401 questions. \smallskip

\noindent \textbf{Qualitative Analysis}: In \figref{fig:qualitative}, we can qualitatively see how our models can reason about relative positions of object and text in the image. Our model picks the correct answer in \figref{fig:qualitative}(a, b, f, g) by reasoning about relations like `\texttt{right}', `\texttt{top-left}'.~\figref{fig:qualitative}(c) shows another examples where our model can reason about spatial relations between object (`green square') and text(`lime'). We can also  see several instances in ~\figref{fig:qualitative}(i, j, k, l) where based on the type of spatial relationship mentioned in the question, our model changes the answer. \smallskip

\noindent \textbf{Potential Sources of Error:} While our model improves multi-modal transformer models by encoding spatial relationships, we are still far away from human baseline. Our models are not robust to spelling mistakes in the OCR tokens~\figref{fig:qualitative}(h). As the models become more visually grounded, they rely on the OCR tokens more often to answer the question. This reduces their ability to learn to pick the right spelling from the static vocabulary in case that word is present. Secondly, these models have trouble generating the stop condition during decoding. As we can see in \figref{fig:qualitative}(d), our model predicts `stop global warming' as the answer whereas the correct answer is `stop.' Finally, while our model can encode relative spatial relationships, for reasoning about absolute positions in the image, our model can benefit from stronger cues about absolute locations. 
\begin{figure*}[t!]
        \begin{center}
        \includegraphics[width=0.85\linewidth]{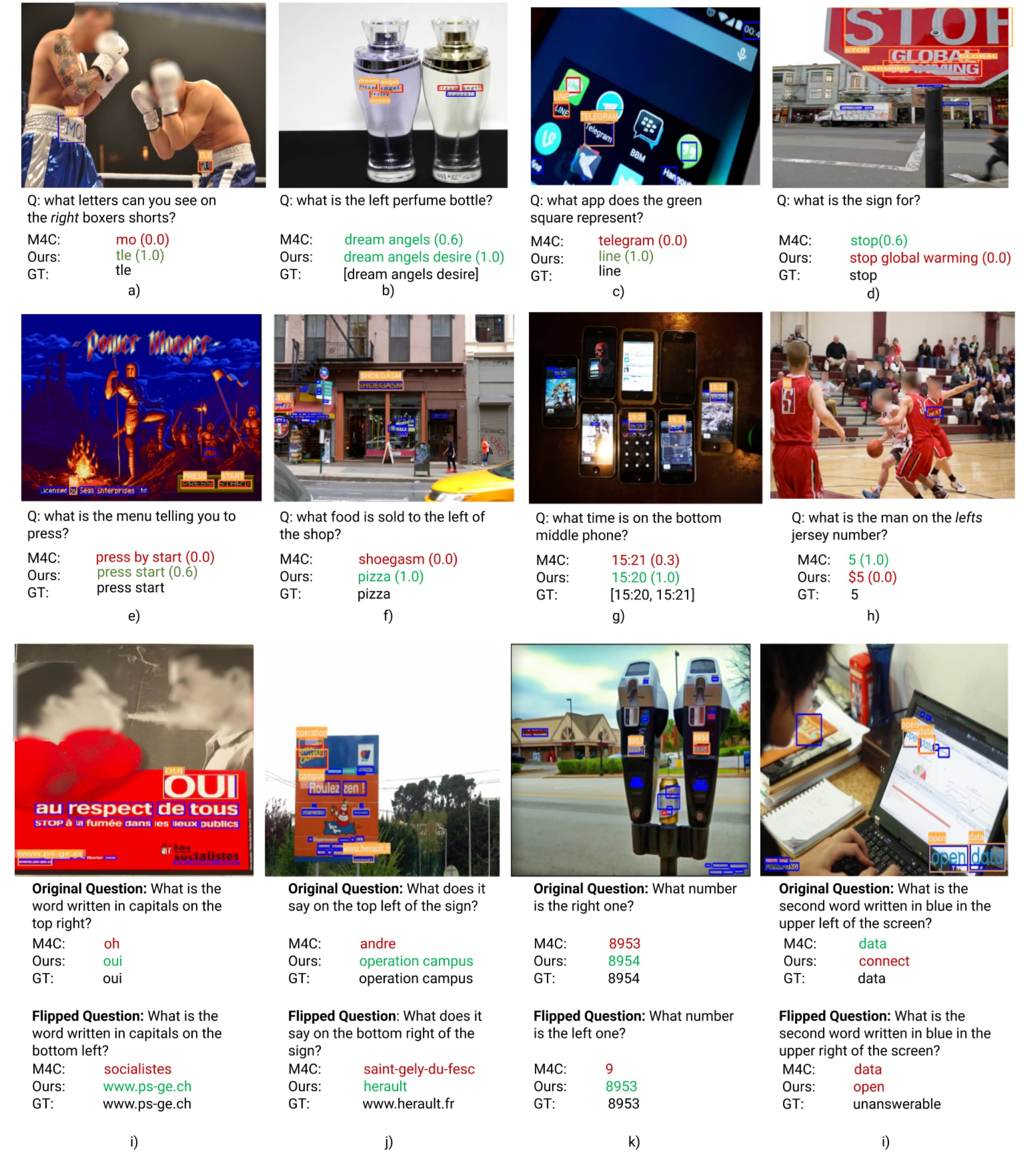}
        \label{fig: nocap_vs_coco_b}
    \end{center}
    \vspace{-25pt}
    \caption{
        \small Qualitative Examples: Top two rows show the output of \mfc and our method on several image-question pairs. The bottom row show examples where we flipped the spatial relation in the original question to see whether the models change their answers.
    }
    \vspace{-18pt}
	\label{fig:qualitative}
\end{figure*}

\vspace{-3pt}
\section{Conclusion}
\vspace{-3pt}
\label{sec:conc}

We developed a spatially aware self-attention layer that encodes different types of relations between input entities via a graph. In our proposed method, each input entity only looks at neighboring entities as defined by a spatial graph. This allows each input to focus on a local context instead of dispersing  attention amongst all other entities. Each head also focuses on a different subset of the spatial relations which avoids learning redundant features. We apply our general framework on the task of TextVQA by constructing a spatial graph between object and OCR tokens and utilizing it in the spatially aware self-attention layers. We found this graph-based attention to significantly improve results achieving state-of-the-art performance on the TextVQA and ST-VQA dataset. Finally, we present our analysis showing how our method improves visual grounding.

\smallskip
\footnotesize \noindent \textbf{Acknowledgements} We thank Abhishek Das and Abhinav Moudgil for their feedback. The Georgia Tech effort was supported in part by NSF, AFRL, DARPA, ONR YIPs, ARO PECASE, Amazon. The views and conclusions contained herein are those of the authors and should not be interpreted as necessarily representing the official policies or endorsements, either expressed or implied, of the U.S. Government, or any sponsor.

{\small
\bibliographystyle{style/ieee_fullname}
\bibliography{strings,main}
}

\clearpage

\title{Appendix}
\author{}
\authorrunning{Y. Kant et al.}
\titlerunning{SA-M4C for TextVQA}
\institute{}
\maketitle

\renewcommand{\thesection}{\Alph{section}}
\renewcommand{\thetable}{\Alph{table}}

\section{Training and Model Parameters:} All the 6-layer models have 96.6 million parameters and the 4-layer models have 82.4 million parameters. We train our models using Adam optimizer~\cite{Kingma2014AdamAM} with a linear warmup and with a learning rate of 1e-4 and a staircase learning rate schedule, where we multiply the learning rate by 0.1 at 14000 and at 19000 iterations. We train for 36.1K total iterations (100 epochs) on 2 NVIDIA Titan XP GPUs for 12 hours and use a batch-size of 96 and $d = 768$ as dimensionality for encoding all multi-modal features. We use the PyTorch  \cite{NEURIPS2019_9015} deep-learning framework for all the experiments.

We list  the hyper-parameters used in our experiments for both \SAM~and \mfc models in Table~\ref{tab:hyperparameters_supp}. We keep these hyper-parameters fixed across all the ablations for both 
TextVQA\cite{Singh2019TowardsVM} and STVQA\cite{STVQA} datasets.
\vspace{-20pt}
\begin{table*}[h!]\scriptsize
\setlength\tabcolsep{3.5pt}
\renewcommand{\arraystretch}{1.25}
\center
\caption{Hyperparameter choices for models.} 
\resizebox{0.8\textwidth}{!}{
\begin{tabular}{clccclc}
\toprule
\# & Hyperparameters & Value & & \# & Hyperparameters & Value\\
\midrule
\band \small\texttt{1} & Maximum question tokens & 20 &&
\small\texttt{2} & Maximum object tokens & 100 \\
\small\texttt{3} & Maximum OCR tokens & 50 &&
\small\texttt{4} & Maximum decoding steps & 12 \\
\band \small\texttt{5} & Embedding size & 768 &&
\small\texttt{6} & Number of Multimodal layers & 6N/2N\textrightarrow4S \\
\small\texttt{7} & Multimodal layer intermediate size & 3072  &&
\small\texttt{8} & Number of attention heads & 12  \\
\band \small\texttt{9} & Types of spatial relationships & 12 &&
\small\texttt{10} & Multimodal layer dropout & 0.1 \\
\small\texttt{11} & Context size & 1/2 &&
\small\texttt{12} & Optimizer & Adam \\
\band \small\texttt{13} & Batch size & 128 &&
\small\texttt{14} & Base Learning rate & 1e-4 \\
\small\texttt{15} & Warm-up learning rate factor & 0.2 &&
\small\texttt{16} & Warm-up iterations & 1000 \\
\band \small\texttt{17} & Vocabulary size & 5000 &&
\small\texttt{18} & Gradient clipping (L-2 Norm) & 0.25 \\
\small\texttt{19} & Number of epochs & 100 &&
\small\texttt{20} & Learning rate decay & 0.1 \\
\band \small\texttt{21} & Learning rate decay steps & 14000, 19000 &&
\small\texttt{22} & Number of iterations & 36000 \\
\midrule
\end{tabular}
}
\smallskip
\label{tab:hyperparameters_supp}
\vspace{-30pt}
\end{table*}

\section{List of spatial-prepositions}
We used the following list of spatial prepositions to form the subset of questions that involve spatial reasoning: north, south, east, west, up, down, left, right, under, top, 
bottom, middle, center, above, below, beside, beneath.
                         
\section{Ablations with varying Spatial Layers}
We also study the affect of using \spat layers in a multimodal transformer. We gradually start replacing the self-attention layers of \mfc with our \spat layers. We  observe from Table~\ref{tab:spatial_supp} that, as we replace more layers, the performance gradually increases. However, it is important to keep a couple of normal self-attention layers at the bottom to allow different modalities to attend to the entire context available to them. Since the \spat layers do not modify the question representations, the self-attention layers in the bottom allow the question tokens to attend to other question tokens as well as object and OCR tokens. Indeed, we see a significant drop as we remove self-attention layers from the bottom.

\vspace{-20pt}
\begin{table*}[h!]\scriptsize
\setlength\tabcolsep{3.5pt}
\renewcommand{\arraystretch}{1.25}
\center
\caption{Ablations with varying number of \spat layers.} 
\resizebox{0.60\textwidth}{!}{
\scriptsize
\begin{tabular}{clccc}
\toprule
& Method & Model Structure & Context & Accuracy on Val. \\
\midrule
\band \texttt{1} & \mfc\cite{hu2019iterative}$^{\dagger}$  & 6N & - & 42.70 \\
\texttt{2} & \SAM & 5N\rarrow1S & 1 & 42.61 \\
\band \texttt{3} & \SAM & 4N\rarrow2S & 1 & 43.19 \\
\texttt{4} & \SAM & 3N\rarrow3S & 1 & 43.16 \\
\band \texttt{5} & \SAM & 2N\rarrow4S & 1 & \textbf{43.80} \\
\texttt{6} & \SAM & 1N\rarrow5S & 1 & 43.07 \\
\midrule
\end{tabular}
}
\smallskip
\label{tab:spatial_supp}
\vspace{-20pt}
\end{table*}

\section{Deforming/Reversing Spatial Graph during Inference}
To understand the role of the spatial graph in our approach, using our best model (\SAM), we experiment by modifying the spatial graph during inference. For this, we reverse every edge type in the spatial graph  (Table~\ref{tab:reverse_supp}, \texttt{Row3}: \SAM~Rev). For instance, the relationship $\langle\text{obj}_{1}-right-\text{obj}_{2}\rangle$ now becomes $\langle\text{obj}_{1}-left-\text{obj}_{2}\rangle$. Similarly, we also experiment by randomly perturbing the spatial graph (Table~\ref{tab:reverse_supp}, \texttt{Row-4}: \SAM~Rand). For this, we replace each existing relationship between two objects with a random one. We observe a significant performance drop in both the experiments which emphasizes  the importance of encoding the spatial relations correctly.

\vspace{-25pt}
\begin{table*}[h!]
\setlength\tabcolsep{3.5pt}
\renewcommand{\arraystretch}{1.25}
\center
\caption{Effect of randomizing and reversing spatial graph during inference.} 
\resizebox{0.73\textwidth}{!}{
\begin{tabular}{clcccccc}
\toprule
& Method & Model Structure & Context & Spatial Graph & \makecell{w/ \\ ST-VQA} & \makecell{Beam \\ size} &
\makecell{Acc. \\ on Val.} \\
\band \small\texttt{1} &  M4C \cite{hu2019iterative}$^{\dagger \dagger}$ & 6N & 2 & - & \cmark & 1 & 43.80 \\
\small\texttt{2} & \SAM \ (ours) & 2N\rarrow4S & 2 & Normal & \cmark & 1 & \textbf{45.10} \\
\midrule
\band \small\texttt{3} & \SAM \ Rev & 2N\rarrow4S & 2 & Reversed & \cmark & 1 & 41.08 \\
\small\texttt{4} & \SAM \ Rand & 2N\rarrow4S & 2 & Randomized & \cmark & 1 & 42.10 \\

\midrule
\end{tabular}
}
\smallskip
\label{tab:reverse_supp}
\vspace{-25pt}
\end{table*}

\noindent \textbf{Performance on questions that involve spatial reasoning}: Additionally, similar to our analysis in the main manuscript, we specifically look at the performance of questions that involve spatial reasoning. On this subset $D_{\text{spa}}$ ($\sim$14\% of the dataset), the performance drops by 4.1\% when the spatial graph is reversed (\SAM~Rev), and drops by 2.6\% when the spatial graph is randomly perturbed (\SAM~Rand). Importantly, on $D_{\text{spa+ocr}}$ which consist of questions that require spatial reasoning and have a majority answer encoded in the OCR tokens, the performance drops drastically by 10\% for \SAM~Rev and 6.6\% for \SAM~Rand.\\

\noindent \textbf{Visual Grounding}: 
As a proxy to analyze visual grounding of our model, we look at instances in which models predict the answer using the list of OCR tokens without relying on the vocabulary. Our model (SA-M4C) picks an answer from the list of OCR tokens for 368/701 questions from the $D_{\text{spa}}$ subset, and achieves 52.85\% accuracy. In contrast, the \SAM~Rev and \SAM~Rand models achieve 39.47\% and 42.43\% accuracy respectively. Similarly, on $D_{\text{spa+ocr}}$ \SAM\ achieves an accuracy of 67.95\%, whereas \SAM~Rev and \SAM~Rand achieve 56.54\% and 52.46\% respectively. \\

In our model, each of the attention heads specializes in encoding a different spatial context. Consequently, we  observe that reversing or randomly changing the spatial context for these heads by deliberate perturbations to the spatial graph has a notable affect on performance. 

\section{Additional Experiments}
\noindent \textbf{ST-VQA Weakly Contextualized Task}: We train SA-M4C with 30k vocabulary and achieve 49.7\% ANLS accuracy beating the previous SoTA by 18.68\% on the Weakly Contextualized Task of ST-VQA. \\  

\noindent \textbf{Adding fully connected heads in the spatial layer}:  We experimented with a model that extends the 12-head spatially-aware layer by adding 6 fully-connected heads that model all spatial relations while keeping the number of parameters comparable to the proposed approach. The performance drops from 43.8\% to 43.41\%.

\clearpage
\section{Qualitative Samples}
\begin{figure*}[h]
        \begin{center}
        \includegraphics[width=1.0\linewidth]{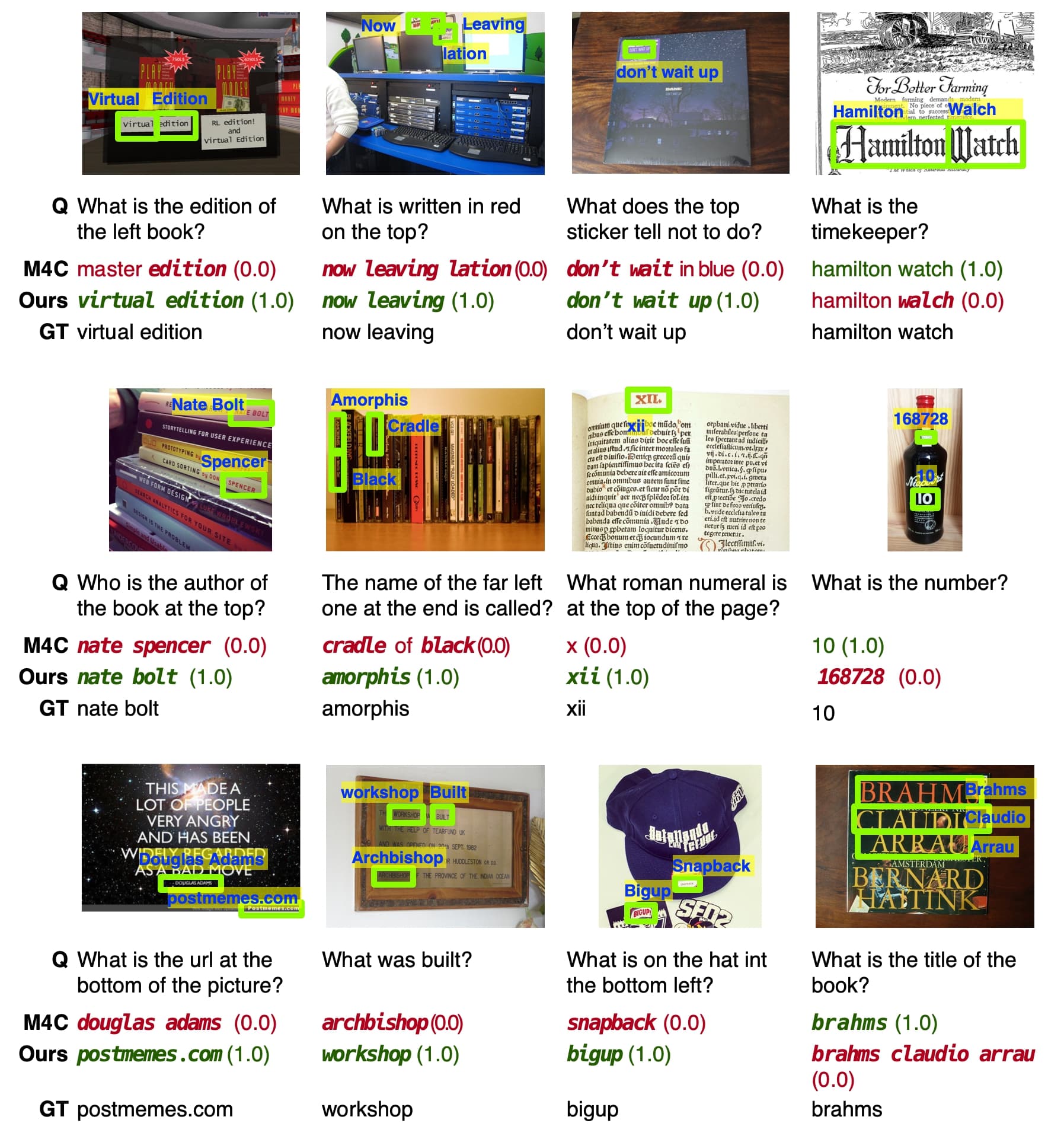}
        \label{fig: nocap_vs_coco_b}
    \end{center}
    \vspace{-15pt}
    \caption{
        Qualitative Examples: The figure shows the output of \mfc and our method on several image-question pairs. \texttt{\textbf{\textit{Bold and italics text}}} denote words chosen from OCR tokens, otherwise it was chosen from the vocabulary. The VQA score for each prediction is mentioned inside parenthesis.
    }
    \vspace{-20pt}
	\label{fig:qualitative}
\end{figure*}

\begin{figure*}[h!]
\vspace{40pt}
        \begin{center}
        \includegraphics[width=1.0\linewidth]{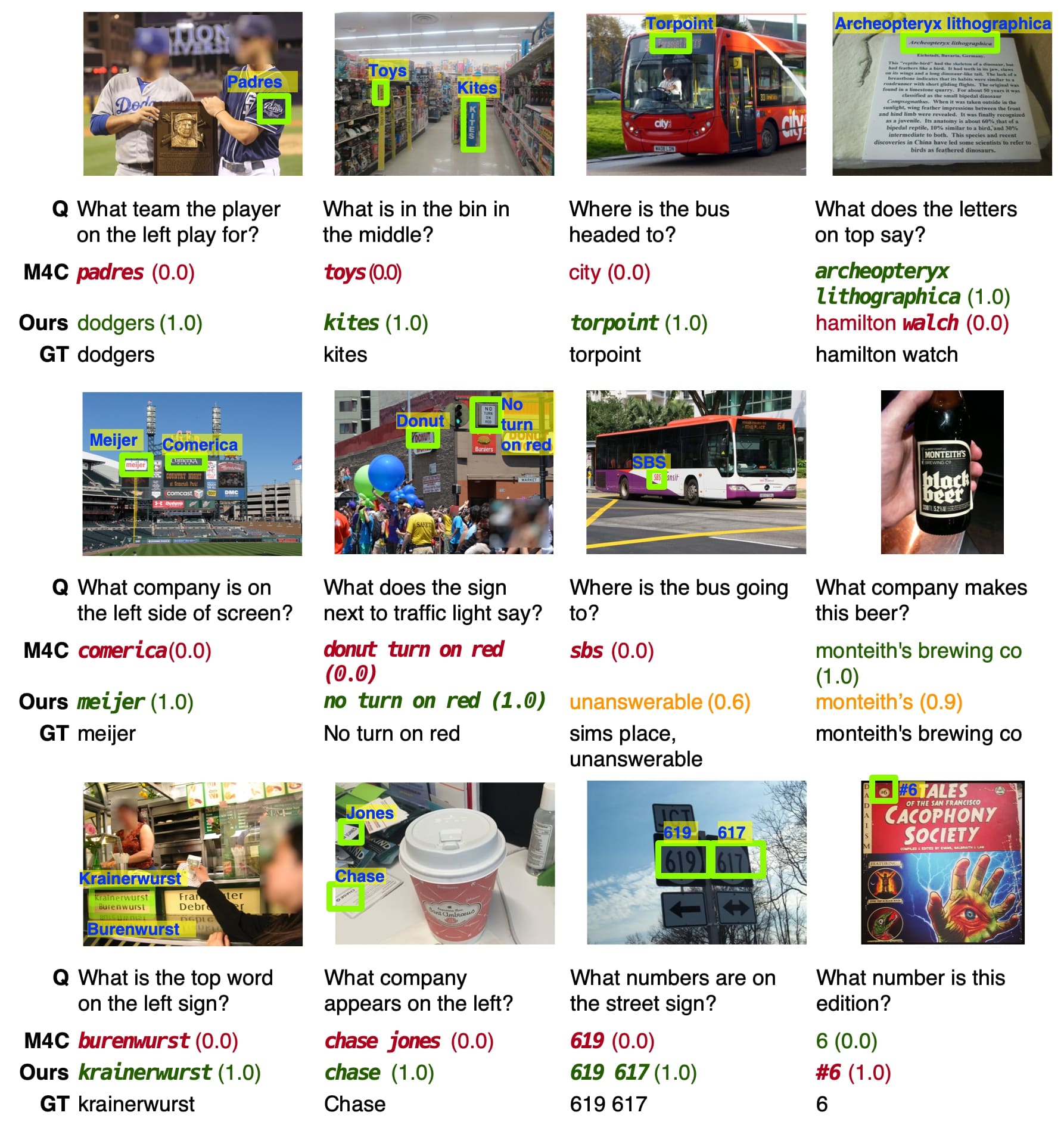}
        \label{fig: nocap_vs_coco_b}
    \end{center}
    \vspace{-15pt}
    \caption{
                Qualitative Examples: The figure shows the output of \mfc and our method on several image-question pairs. \texttt{\textbf{\textit{Bold and italics text}}} denote words chosen from OCR tokens, otherwise it was chosen from the vocabulary. The VQA score for each prediction is mentioned with parenthesis.
    }
    \vspace{50pt}
	\label{fig:qualitative}
\end{figure*}

\begin{figure*}[t!]
    \vspace{30pt}
        \begin{center}
        \includegraphics[width=1.0\linewidth]{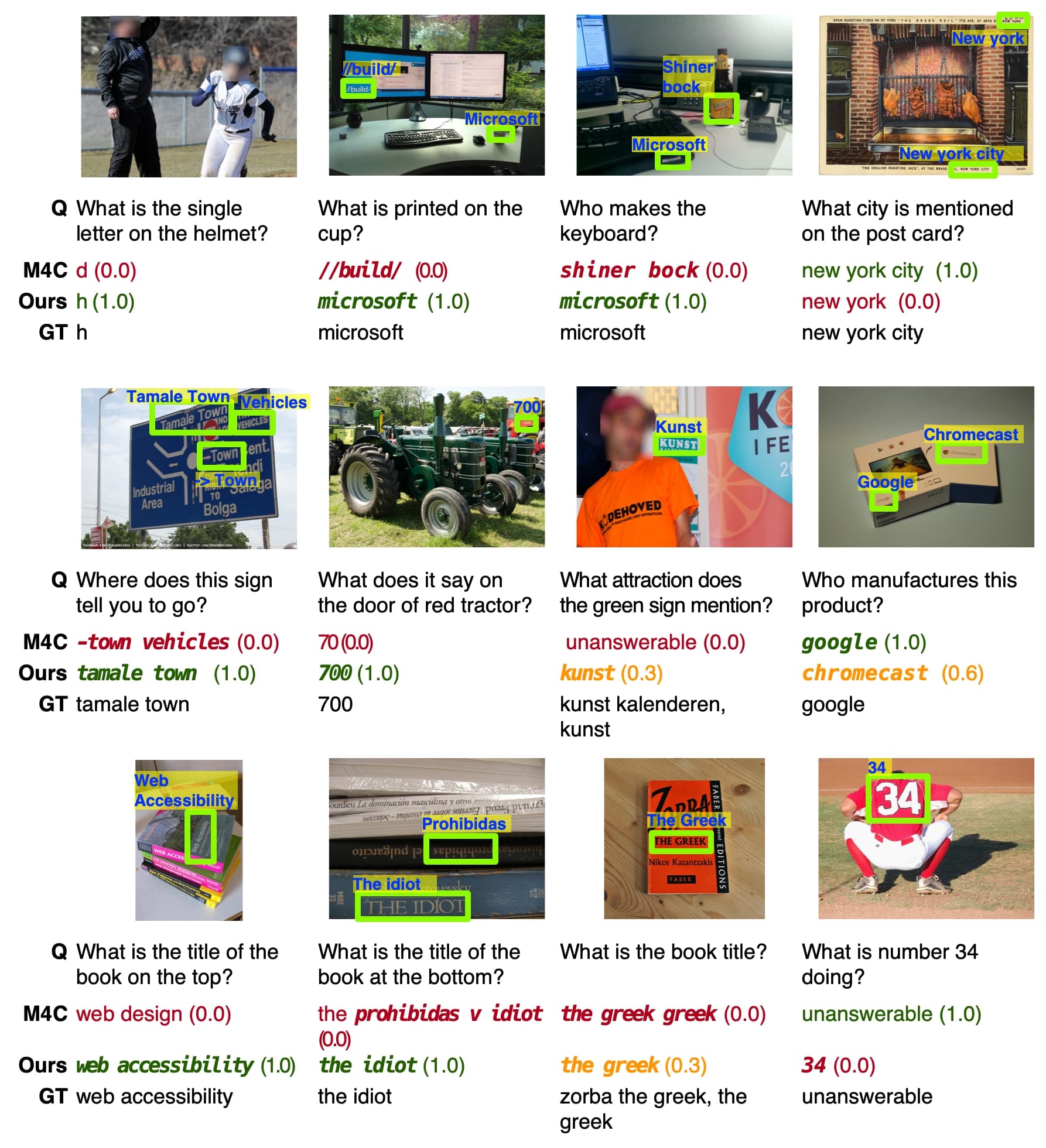}
        \label{fig: nocap_vs_coco_b}
    \end{center}
    \vspace{-15pt}
    \caption{
               Qualitative Examples: The figure shows the output of \mfc and our method on several image-question pairs. \texttt{\textbf{\textit{Bold and italics text}}} denote words chosen from OCR tokens, otherwise it was chosen from the vocabulary. The VQA score for each prediction is mentioned with parenthesis.
    }
    \vspace{50pt}
	\label{fig:qualitative}
\end{figure*}

\definecolor{orange}{rgb}{1,0.57,0}
\definecolor{red}{rgb}{0.69,0.0,0.10}
\definecolor{green}{rgb}{0.13,0.36,0.00}
\newcommand{\correct}[1]{\textcolor{green}{#1}}
\newcommand{\parcor}[1]{\textcolor{orange}{#1}}
\newcommand{\wrong}[1]{\textcolor{red}{#1}}

\begin{figure*}[t!]
        \begin{center}
        \includegraphics[width=1.0\linewidth]{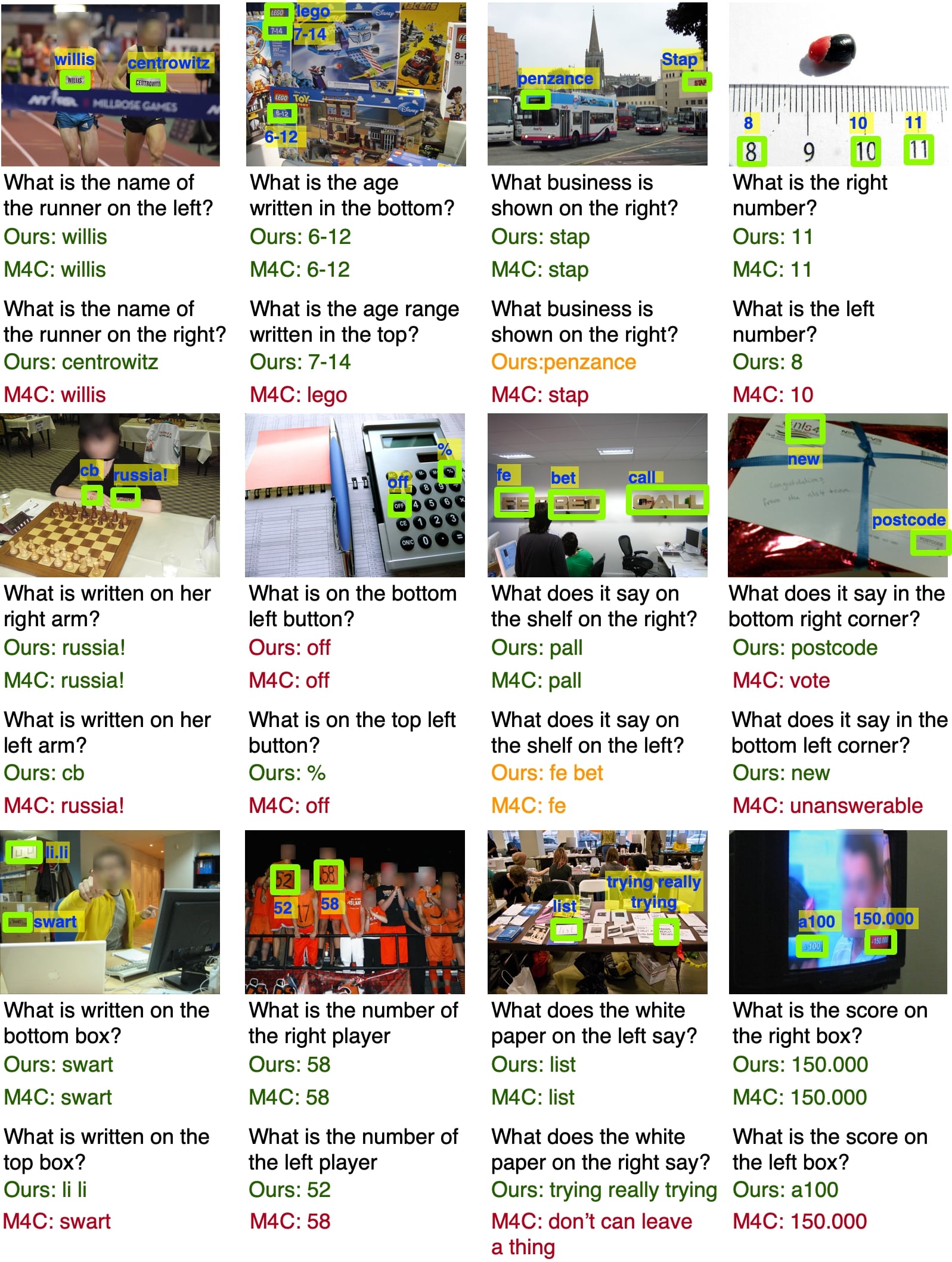}
        \label{fig: nocap_vs_coco_b}
    \end{center}
    \vspace{-15pt}
    \caption{
        The figure shows examples where we flipped the spatial relation in the original question to see whether the models change their answers. We observe that our spatially aware multimodal transformer correctly reasons about the  spatial relationships mentioned in the question and predict the answer more accurately than \mfc. \correct{Green} text denote correct predictions. \wrong{Red} text denote incorrect predictions while \parcor{orange} text denote partially correct answers.
    }
    \vspace{20pt}
	\label{fig:qualitative}
\end{figure*}

\end{document}